\documentclass{article}

\usepackage[preprint,nonatbib]{neurips_2025}

\usepackage[utf8]{inputenc} % allow utf-8 input
\usepackage[T1]{fontenc}    % use 8-bit T1 fonts
\usepackage{hyperref}       % hyperlinks
\usepackage{url}            % simple URL typesetting
\usepackage{booktabs}       % professional-quality tables
\usepackage{amsfonts}       % blackboard math symbols
\usepackage{nicefrac}       % compact symbols for 1/2, etc.
\usepackage{microtype}      % microtypography
\usepackage{xcolor}         % colors 
\usepackage{graphicx}
\usepackage{subcaption}
\usepackage{adjustbox}
\usepackage[accsupp]{axessibility} 
\usepackage{hyperref}
\usepackage{orcidlink}
\usepackage{amsmath}
\usepackage{amssymb}
\usepackage{mathtools}
\usepackage{makecell}
\usepackage{multirow}
\usepackage{siunitx}
\usepackage{colortbl}
\usepackage{enumitem}
\usepackage{rotating}

\title{Omni-WorldBench: Towards a Comprehensive Interaction-Centric Evaluation for World Models}

\author{%
  \textbf{Meiqi Wu}{$^\ddagger$}\textsuperscript{\rm1,\rm2,\rm5},
  \textbf{Zhixin Cai}{$^\ddagger$}\textsuperscript{\rm3,\rm5},
  \textbf{Fufangchen Zhao}{$^\ddagger$}\textsuperscript{\rm4,\rm5},
  \textbf{Xiaokun Feng}\textsuperscript{\rm2},
  \textbf{Rujing Dang}\textsuperscript{\rm5}, \\
  \textbf{Bingze Song}\textsuperscript{\rm5},
  \textbf{Ruitian Tian}\textsuperscript{\rm5},
  \textbf{Jiashu Zhu}\textsuperscript{\rm5},
  \textbf{Jiachen Lei}\textsuperscript{\rm5},
  \textbf{Hao Dou}\textsuperscript{\rm5},
  \textbf{Jing Tang}\textsuperscript{\rm5}, \\
  \textbf{Lei Sun}\textsuperscript{\rm5},
  \textbf{Jiahong Wu}{$^*$}\textsuperscript{\rm5},
  \textbf{Xiangxiang Chu}\textsuperscript{\rm5},
  \textbf{Zeming Liu}\textsuperscript{\rm3},
  \textbf{Kaiqi Huang}{$^*$}\textsuperscript{\rm2}
  \\
  \textsuperscript{\rm1}School of Computer Science and Technology, UCAS \\
  \textsuperscript{\rm2}The Key Laboratory of Cognition and Decision Intelligence for Complex Systems, CASIA \\
  \textsuperscript{\rm3}School of Computer Science and Engineering, Beihang University \\
  \textsuperscript{\rm4}State Key Laboratory of Networking and Switching Technology, BUPT \\
  \textsuperscript{\rm5}AMAP, Alibaba Group
\\
}

\PassOptionsToPackage{numbers, compress}{natbib}
\usepackage{amssymb}
\usepackage{multirow}

\begin{document}
% 统计，使用斜体\emph{}, 使其居中\begin{center}，新的一段\paragraph{}

\maketitle
\def\thefootnote{$\ddagger$}\footnotetext{Work done during the internship at AMAP, Alibaba Group.}
\def\thefootnote{$*$}\footnotetext{Corresponding author.}
\begin{abstract}

Video-based world models have emerged along two dominant paradigms: video generation and 3D reconstruction. However, existing evaluation benchmarks either focus narrowly on visual fidelity and text–video alignment for generative models, or rely on static 3D reconstruction metrics that fundamentally neglect temporal dynamics. We argue that the future of world modeling lies in 4D generation, which jointly models spatial structure and temporal evolution. In this paradigm, \textbf{the core capability is interactive response}: the ability to faithfully reflect how interaction actions drive state transitions across space and time. Yet no existing benchmark systematically evaluates this critical dimension.
To address this gap, we propose Omni-WorldBench, a comprehensive benchmark specifically designed to evaluate the interactive response capabilities of world models in 4D settings. Omni-WorldBench comprises two key components: Omni-WorldSuite, a systematic prompt suite spanning diverse interaction levels and scene types; and Omni-Metrics, an agent-based evaluation framework that quantifies world modeling capabilities by measuring the causal impact of interaction actions on both final outcomes and intermediate state evolution trajectories. We conduct extensive evaluations of 18 representative world models across multiple paradigms. Our analysis reveals critical limitations of current world models in interactive response, providing actionable insights for future research. Omni-WorldBench will be publicly released to foster progress in interactive 4D world modeling.

\end{abstract}

\begin{figure}[!h]
    \centering
    \includegraphics[width=\linewidth]{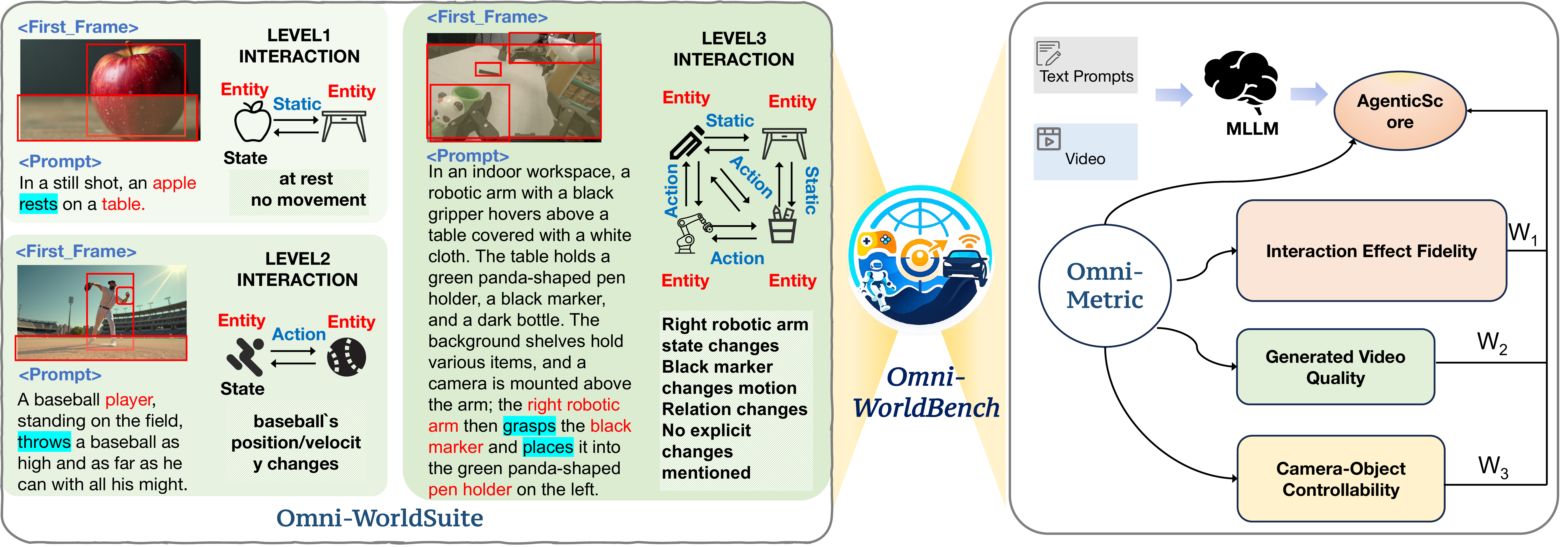}
    \caption{Overview of Omni-WorldBench. \textbf{Left:} Omni-WorldSuite defines three levels of interactions, each specified by an initial frame and a prompt. \textbf{Right:} Omni-Metrics comprises an evaluation pipeline that measures interaction effect fidelity, generated video quality, camera-object controllability, and spatiotemporal causal coherence, and then employs an MLLM to adaptively fuse these signals into the final AgenticScore.
    }
\label{fig:intro}
\end{figure}

\section{Introduction}
\label{sec:intro}

%%%% 0301 fxk_V2
The world models aim to characterize the temporal evolution of environmental states under given interaction conditions, providing a foundation for counterfactual reasoning, planning, and decision-making \cite{ha2018world}. Taking advantage of recent advances in video generation, this paradigm has increasingly adopted video synthesis as a core implementation pathway. By leveraging high-quality general-purpose video representations to model world dynamics, video-based world models have been widely applied to autonomous driving, embodied intelligence, and game agents, substantially accelerating progress in these domains.

Unlike rapid progress in world model design, the development of dedicated evaluation benchmarks appears to be somewhat lagging. Existing evaluation methods largely rely on conventional video generation metrics, such as FID and FVD, or adopt general-purpose evaluation benchmarks (e.g., VBench \cite{huang2024vbench}). 
Although these metrics are effective in measuring visual fidelity and text–video alignment \cite{liu2024survey}, 
they do not adequately capture the core capability of world models—the ability to generate consistent and plausible responses under varying interaction conditions.

To comprehensively evaluate the interactive response capabilities of world models, we propose a novel benchmark, \textbf{Omni-WorldBench} (Fig.~\ref{fig:intro}).
At its core, we construct a systematic prompt suite, \textbf{Omni-WorldSuite}, designed to thoroughly assess model performance across diverse interaction levels and scenario types.
Specifically, interaction conditions can produce effects confined to a single object, extend to the local environment, or induce global environmental changes. These progressively increasing interaction scopes impose distinct representational and dynamic modeling requirements on world models. Consequently, the evaluation prompts in Omni-WorldSuite are systematically organized around these three hierarchical interaction levels.
Furthermore, since world modeling is a broad and application-dependent research paradigm, existing studies are often grounded in specific domains such as autonomous driving, embodied robotics, and gaming environments. 
To ensure that Omni-WorldSuite is applicable to both general-purpose video generation models and scenario-specific world models, our evaluation prompts also encompass real-world physical settings as well as representative application domains.

To complement Omni-WorldSuite, we establish a comprehensive and effective evaluation protocol, \textbf{Omni-Metric}, designed to holistically assess the fidelity and consistency of world state representations. Distinct from prior works that predominantly focus on static visual fidelity \cite{duan2025worldscore, huang2024vbench++}, Omni-Metrics explicitly extends the evaluation toward dynamic, controllable, and interaction-aware generation, which are essential to world models. Specifically, Omni-Metrics evaluates models from three complementary aspects. First, \textit{Generated Video Quality} extends evaluation beyond static appearance to dynamic perceptual quality, measuring temporal flickering, motion smoothness, content alignment, and dynamic degree to capture the visual coherence of generated sequences over time. Second, \textit{Camera-Object Controllability} assesses whether a model can follow explicit camera instructions while maintaining coherent object behavior, and further evaluates long-horizon continuity through a novel scene transition metric, \textit{Transitions Detect}. Third, \textit{Interaction Effect Fidelity} targets the core capability of interactive world modeling by examining whether actions induce the expected effects on intervened objects in a physically plausible and causally consistent manner, supported by quantitative indicators of action-effect correspondence, physical principles, and spatial logic. Since these dimensions are heterogeneous yet complementary, we further introduce an agent-based aggregation framework that adaptively combines outputs from multiple evaluation tools according to prompt semantics, yielding a unified overall metric, \textbf{AgenticScore}, for more reliable evaluation.

Finally, we conduct a systematic evaluation of 18 representative world models, and the results comprehensively reveal the performance boundaries and limitations of current models in interactive response capabilities.
Further human alignment studies demonstrate that Omni-Metric  aligns well with human preferences, validating its effectiveness in assessing world model performance.
Our key contributions are as follows: 
\begin{enumerate}
\item To address the critical absence of standardized evaluation protocols, we introduce \textbf{Omni-WorldBench}. To the best of our knowledge, this is the \textbf{first} benchmark dedicated to assessing the interactive response capabilities of world models, offering a \textbf{comprehensive and holistic} evaluation framework rather than a narrow capability test.
\item We establish a rigorous evaluation infrastructure comprising \textbf{Omni-WorldSuite}, a hierarchical prompt suite spanning diverse interaction levels and scenario types, and \textbf{Omni-Metric}, an agent-based evaluation protocol that quantitatively measures the impact of actions on both final outcomes and intermediate state transitions.
\item We conduct a comprehensive evaluation of 18 video generation models and world models, systematically analyzing their performance. Our findings unveil critical limitations in the 4D interactivity capabilities of current world models, highlighting key areas for improvement. Additionally, we propose a curated benchmark, offering to guide and accelerate future advancements in 4D world model generation.
% \item We conduct a comprehensive evaluation of 18 video generation models and world models, systematically analyzing their performance. Our findings unveil critical limitations in the 4D interactivity capabilities of current world models, highlighting key areas for improvement. Additionally, we propose a curated benchmark and human-alignment evaluation metrics, offering to guide and accelerate future advancements in 4D world model generation.
\end{enumerate}

\section{Related Works}

\subsection{World Models Design}
World models characterize how environment states evolve over time under given interaction conditions, thereby providing effective support for tasks such as counterfactual simulation, planning, and decision-making \cite{ha2018world}.
Early world models primarily relied on multimodal large language models (MLLMs) \cite{hurst2024gpt,Qwen3-VL,bai2025univg,chu2025gpg} that represent world states through textual abstractions \cite{yang2024rila,shridhar2020alfworld}.  Recent advances in video generation \cite{sora_initial,wan2025,mao2025omni,wu2026latent,zhu2026artifact} have driven a shift toward video-based world models, which offer a more expressive and grounded representation of complex environments and have emerged as a dominant paradigm in the field \cite{ding2025understanding,zhu2024sora,zheng2026code2world}. In this work, we focus on world models built upon video generation.

Across different application domains, video-based world models have followed distinct yet intrinsically related technical trajectories. In autonomous driving, world models primarily focus on long-horizon traffic scene evolution and the decision-making of vehicle agents \cite{feng2025survey}. Representative works such as GAIA \cite{hu2023gaia}, DriveDreamer \cite{wang2024drivedreamer}, DrivingWorld \cite{hu2024drivingworld}, and Vista \cite{gao2024vista} leverage action-conditioned future frame prediction to support planning and simulation. 
In embodied intelligence and robotics, world models place greater emphasis on object-centric dynamics and manipulation control \cite{long2025survey}. Methods such as IRASim \cite{zhu2024irasim}, Cosmos \cite{agarwal2025Cosmos}, RoboScape \cite{shang2025roboscape} and LVP \cite{chen2025large} tightly integrate perception, action, and physical reasoning to simulate interaction-driven environment changes. 
In game environments, works including Genie \cite{bruce2024genie,parker2024genie}, Matrix-Game \cite{zhang2025matrix,he2025matrix}, WorldPlay \cite{sun2025worldplay}, and Hunyuan-GameCraft \cite{li2025hunyuan,tang2025hunyuan} aim to construct highly interactive and playable virtual worlds. Despite differences in input modalities, action spaces, and domain-specific constraints, these methods share a common objective: learning how the environment responds coherently to different interaction instructions. This highlights interaction as a core capability of world modeling \cite{ding2025understanding,agarwal2025Cosmos}.
Motivated by this, our benchmark takes interaction as the central axis for evaluating world models.

\subsection{World Models Evaluation}

Despite the rapid progress of video-based world models, the development of corresponding evaluation benchmarks has remained relatively limited \cite{ding2025understanding}. Early studies \cite{feng2024matrix,yu2025gamefactory,guo2025mineworld,chen2025finger,li2025next} primarily rely on generic metrics, such as FID \cite{heusel2017gans}, IS \cite{salimans2016improved}, FVD \cite{unterthiner2019fvd}, which often exhibit significant deviations from human perceptual judgments \cite{ding2022cogview2,otani2023toward, ling2025vmbench, wu2026imagerysearch}.
Subsequently, several evaluation tools originally designed for video generation, such as VBench \cite{huang2024vbench}, have been introduced \cite{che2024gamegen,tang2025hunyuan,ling2025vmbench,feng2025narrlv,zhu2026artifact}. While these benchmarks play an important role in assessing overall visual quality and text–video alignment \cite{liu2024survey,huang2024mmgenbench}, they struggle to adequately characterize the core interactive capabilities of world model tasks. As a result, such metrics provide only limited insights for the design and analysis of interactive world models.
Moreover, WorldScore \cite{duan2025worldscore} has been proposed as a benchmark specifically tailored to world models. It focuses on evaluating a model’s ability to generate geometrically consistent 3D scenes under viewpoint changes, emphasizing spatial coherence and geometric realism. Although this represents an important step toward world-model-aware evaluation, the considered form of interaction is largely restricted to camera motion. In contrast, contemporary world models increasingly emphasize a broader range of interaction types \cite{tang2025hunyuan,chen2025large}.
Motivated by this gap, we introduce Omni-WorldBench, an interaction-centric evaluation benchmark that systematically covers multiple levels of interaction complexity. We hope that Omni-WorldBench can serve as a comprehensive tool for characterizing the interactive expressiveness of world models.

\section{Omni-WorldSuite}
\label{sec:interworld_suite}

To enable a comprehensive analysis of the interactive response capabilities of world models, \textbf{Omni-WorldSuite} constructs targeted evaluation prompts across diverse interaction levels and scenario types. In this section, we detail the construction pipeline of Omni-WorldSuite, provide representative examples, and present its statistical analysis.

\subsection{Construction Pipeline}

% The prompts in \textbf{Omni-WorldSuite} are constructed along two primary dimensions. 
% First, to capture interaction modeling—the core capability of world models—we categorize interactions by their scope of influence, namely a single object, a local region, or the global environment, forming three interaction levels. 
% Second, to ensure applicability to both general-purpose video generation models and domain-specific world models, the prompts cover both real-world general scenes and representative task-oriented scenes, such as autonomous driving, embodied robotics, and gaming environments. 
% By default, each evaluation prompt consists of a textual prompt describing the world evolution process and an initial frame image specifying the starting world state. Fig.~\ref{fig:promptsuite-pipeline} (a) and (b) illustrate two strategies for constructing the evaluation prompts.

\begin{figure}
    \centering
    \includegraphics[width=\linewidth]{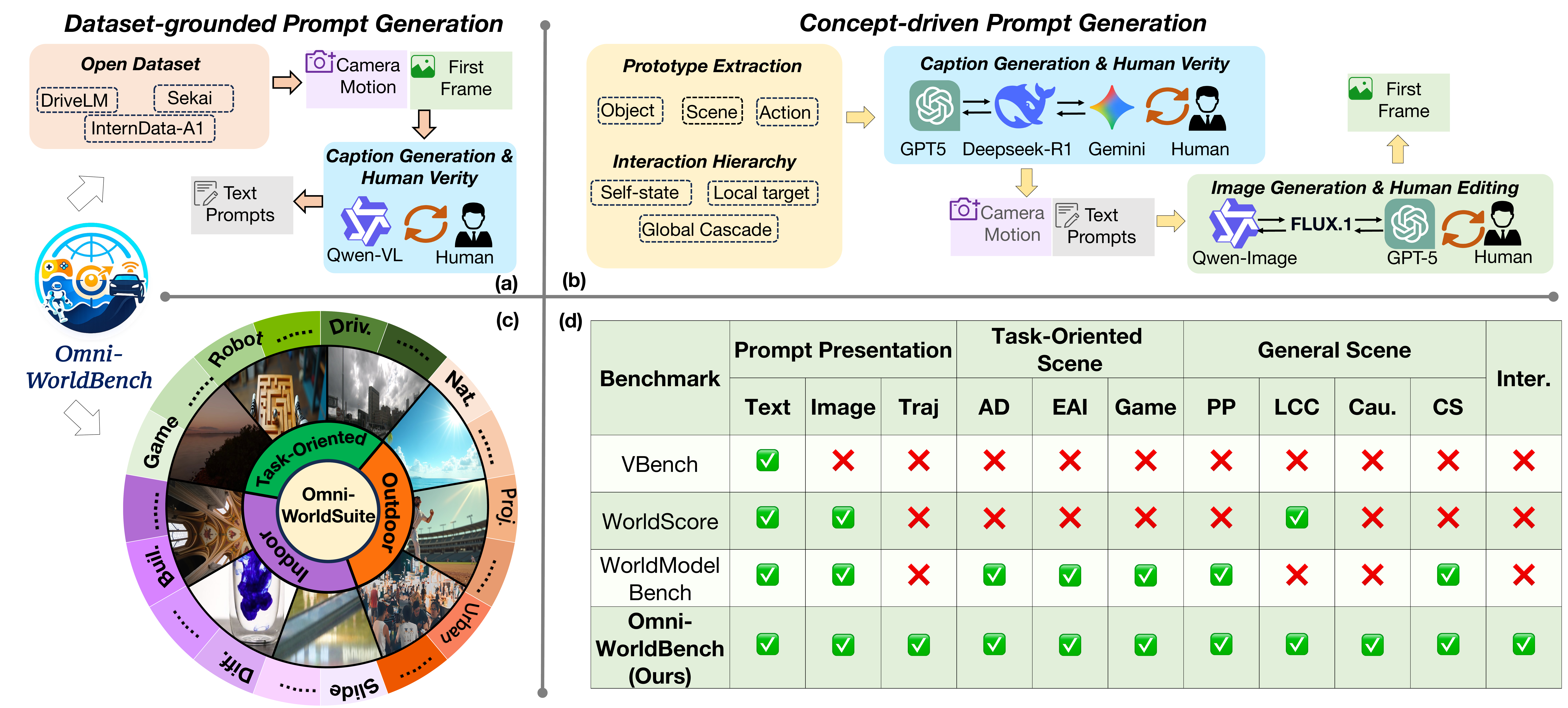}
    \caption{Omni-WorldSuite Construction Pipeline and Analysis. \textbf{(a)} Dataset-grounded prompt generation. Prompts are generated from open-source data using first-frame and camera-motion cues, refined through VLM captioning, and finally verified by human annotators. \textbf{(b)} Concept-driven prompt generation. Prompts are derived from interaction prototypes using LLM/VLM-based generation and human curation, together with generated or edited first frames. \textbf{(c)} Suite taxonomy across indoor scenes, including diffusion (Diff.), sliding, and building-related (Buil.) scenarios; outdoor scenes, including natural, projectile motion (Proj.), and urban scenarios (Urban); and task-oriented settings, including robotics (Robot), autonomous driving (Driv.), and gaming (Game). \textbf{(d)} Coverage comparison by prompt modality and capability axes. Abbr.: \textbf{Traj} (camera trajectory); \textbf{AD} (autonomous driving), \textbf{EAI} (embodied AI); \textbf{PP} (physical principles); \textbf{LCC} (loop-closure consistency); \textbf{Cau.} (causality); \textbf{CS} (common sense); \textbf{Inter.} (Interaction).}
    \label{fig:promptsuite-pipeline}
\end{figure}

The prompts in \textbf{Omni-WorldSuite} are designed along two primary dimensions. The first dimension is scene coverage, spanning both general daily-life scenarios and task-oriented environments such as autonomous driving, embodied AI, and gaming. Collectively, these scenarios cover key aspects of world modeling, including physical laws, commonsense reasoning, causality, camera motion, closed-loop dynamics, and spatial constraints. The second dimension is a three-level interaction hierarchy that characterizes the scope of interaction effects (Fig.~\ref{fig:intro} (Left)). Level 1 involves actions whose effects are confined to the acting object, without altering other objects or the surrounding environment. Level 2 includes localized interactions where one object directly affects another. Level 3 captures more complex interactions that influence multiple objects and lead to broader environmental changes. Each prompt is defined by a textual description of interaction-driven world-state evolution and an initial frame image specifying the starting world state. For a subset of prompts that require explicit camera control, we additionally provide camera trajectories to constrain the viewpoint transition during generation. 
Fig.~\ref{fig:promptsuite-pipeline}(a) and (b) illustrate two prompt construction strategies.

\paragraph{Dataset-grounded Prompt Generation.} As shown in Fig.~\ref{fig:promptsuite-pipeline}(a), we introduce a dataset-grounded prompt construction strategy to address the limited realism, complexity, and robustness of synthetic images. We first extract the camera motion trajectory and the first video frame from open-source datasets to serve as the motion and visual prompts, respectively. Next, we employ Qwen-VL~\cite{Qwen3-VL} to generate an initial caption for the sequence. To mitigate potential errors in spatial relations and object attributes, all generated captions are manually verified and refined to ensure consistency with the source sequence. The final evaluation prompt consists of the verified caption, the initial frame, and, when available, the original camera trajectory, serving as the grounded input for benchmark evaluation. Specifically, Omni-WorldSuite covers three domains: 
\begin{itemize}
    \item \textit{Autonomous Driving}, which uses sequences from DriveLM~\cite{sima2024drivelm}. We extract the first-frame ego-view image together with recorded camera trajectories to evaluate the model's ability to extrapolate road dynamics under realistic driving conditions. 
    \item \textit{Embodied Robotics}, which uses manipulation-oriented tasks from InternData-A1~\cite{cai2026internvla} to examine physical causality arising from robot--object interactions. 
    \item \textit{Gaming and Simulation}, which uses Sekai~\cite{li2025sekai} to test whether the model can preserve coherent motion patterns in highly dynamic and non-photorealistic environments. 
\end{itemize}

\paragraph{Concept-driven Prompt Generation.} As shown in Fig.~\ref{fig:promptsuite-pipeline}(b), we introduce a concept-driven prompt construction strategy featuring a generate--verify--refine pipeline to synthesize text, first frames (representing the initial world state), and camera motion trajectories.
Specifically, we first \textit{build a set of prototype concepts} spanning scene domains, target objects, and actions under different interaction levels. 
We then randomly sample an interaction level, scene type, target entity, and action from the resulting taxonomy. Conditioned on these attributes, ChatGPT-5.2~\cite{choi2025chatgpt} \textit{generates a textual prompt and a camera trajectory}. Both outputs are further cross-checked by Gemini~\cite{google2025gemini3} and DeepSeek-R1~\cite{guo2025deepseek}, followed by careful human verification and refinement. This manual revision process eliminates linguistic ambiguity and ensures the clarity, motion plausibility, and overall consistency of the evaluation cases.
Finally, we adopt a \textit{multi-stage image generation} pipeline to ensure high-fidelity initial frames. We use FLUX.1-dev~\cite{flux1kreadev2025} to generate $3$ candidates per prompt with a CFG scale of $3.5$ and $50$ sampling steps. All candidates are manually screened for physical plausibility, instruction adherence, and visual quality. If no valid result is obtained, we rewrite the prompt with ChatGPT-5.2 and, when necessary, apply Qwen-Image~\cite{wu2025qwenimagetechnicalreport} for refinement or artifact correction. Only minor localized in-painting is allowed during post-processing. All final images must satisfy quality control requirements, including a minimum resolution of $1024 \times 1024$, consistency with the prompt, and clear visibility of the target interactive objects.

\begin{figure}
    \centering
    \includegraphics[width=\linewidth]{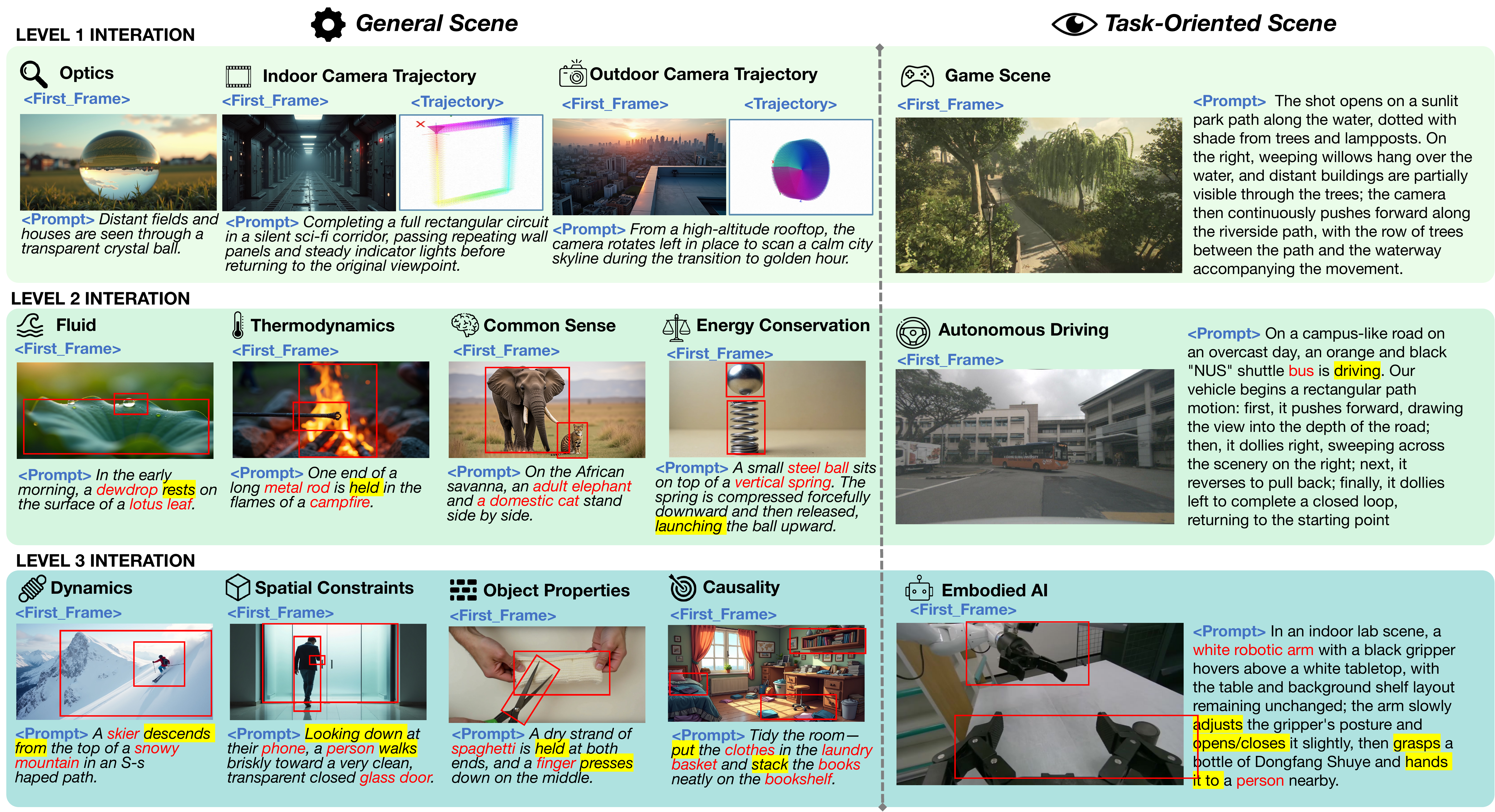}
  \caption{\textbf{Omni-WorldSuite examples across three interaction levels.} \textbf{Left:} Examples from the \textit{General Scene} domain. \textbf{Right:} Examples from the \textit{Task-Oriented Scene} domain, including optics/camera trajectories, game, physics/common sense, autonomous driving, and embodied AI. Each example pairs a first-frame grounding (and trajectory, when applicable) with an action prompt, with red boxes indicating interaction-relevant entities.}
\label{fig:dataset-examples}
\end{figure}
\paragraph{Omni-WorldSuite Examples.}
As Fig.~\ref{fig:dataset-examples} illustrates, we pair initial frames with action-driven prompts to demonstrate the three-level interaction hierarchy, visually anchoring relevant entities with red boxes.
\begin{itemize}
    \item \textbf{Level 1:} Actions are confined to the acting object without altering other objects or the environment. \textit{General Scenes} evaluate phenomena like physical optics (e.g., viewing fields through a crystal ball), while \textit{Task-Oriented Scenes} test continuous spatial navigation (e.g., moving along a riverside path).
    \item \textbf{Level 2:} One object directly affects another. Examples include testing thermodynamics in \textit{General Scenes} (e.g., heating a metal rod in a campfire) and complex ego-vehicle navigation alongside dynamic traffic in \textit{Task-Oriented Scenes} (e.g., autonomous driving).
    \item \textbf{Level 3:} Actions influence multiple objects and lead to broader environmental changes. Prompts cover physical causality in \textit{General Scenes} (e.g., snapping spaghetti, tidying a room) and multi-stage manipulation in \textit{Task-Oriented Scenes} (e.g., a robotic arm grasping a bottle and handing it to a person).
\end{itemize}

\subsection{Omni-WorldSuite Analysis and Statistics}

\paragraph{Concept Set Analysis.}
As shown in Fig.~\ref{fig:promptsuite-pipeline}(c), the set of prototype concepts mainly covers two broad scene categories, namely indoor and outdoor scenes, as well as task-oriented scenarios such as autonomous driving, embodied robotics, and gaming. Within each broad category, we further include several representative interaction types. Overall, these prompts span multiple dimensions, ranging from natural environments, urban scenes, and architectural spaces to fundamental physical motion, fluid and thermal phenomena, optical effects, material deformation, commonsense reasoning, object affordance, robotic manipulation, and embodied interaction, thereby forming a comprehensive prompt set that balances scene diversity, physical realism, and task interactivity. Beyond static scene descriptions, the collection also includes a large number of dynamic processes, causally driven events, and goal-oriented manipulation tasks, enabling a systematic evaluation of a model’s capabilities in scene understanding, physical consistency, spatial constraint reasoning, and embodied task execution.

To facilitate the computation of evaluation metrics, we further provide auxiliary metadata for each prompt. (i) First, we enumerate all entity objects appearing in the prompt and categorize them into affected and unaffected sets according to the interaction actions. For affected entities, we additionally annotate the expected coarse motion direction and magnitude. (ii) Next, based on the world evolution described in the textual prompt, we extract a list of key events ordered by their temporal occurrence. (iii) Finally, to evaluate camera motion and spatial consistency, we annotate expected camera motions for a subset of prompts, including the motion direction and magnitude. We also incorporate a challenging return-to-origin setting, where the model is required to return the camera to its original position after completing a motion cycle. Video frames in which the camera revisits the same spatial position are referred to as revisit frames.

\paragraph{Compare with other Benchmarks.}
As shown in Fig.~\ref{fig:promptsuite-pipeline}(d), compared with prior benchmarks such as VBench~\cite{zheng2025vbench2}, WorldScore~\cite{duan2025worldscore}, and WorldModelBench~\cite{li2025worldmodelbench}, Omni-WorldBench supports the most comprehensive set of prompt modalities, encompassing text, image, and trajectory inputs. Moreover, it evaluates both task-oriented and general scenes, rather than focusing on only a narrow subset of scenarios. Specifically, it covers a diverse range of scene and reasoning types, including physical regularities, loop-closure motion, causal reasoning, and commonsense reasoning, thereby achieving the broadest coverage of scenario types among existing benchmarks. Furthermore, Omni-WorldBench is the first benchmark to explicitly account for interaction types as a core evaluation dimension. This comprehensive design provides a reliable testbed for the development and evaluation of next-generation 4D world models.

\begin{figure}
    \centering
    \includegraphics[width=1\linewidth]{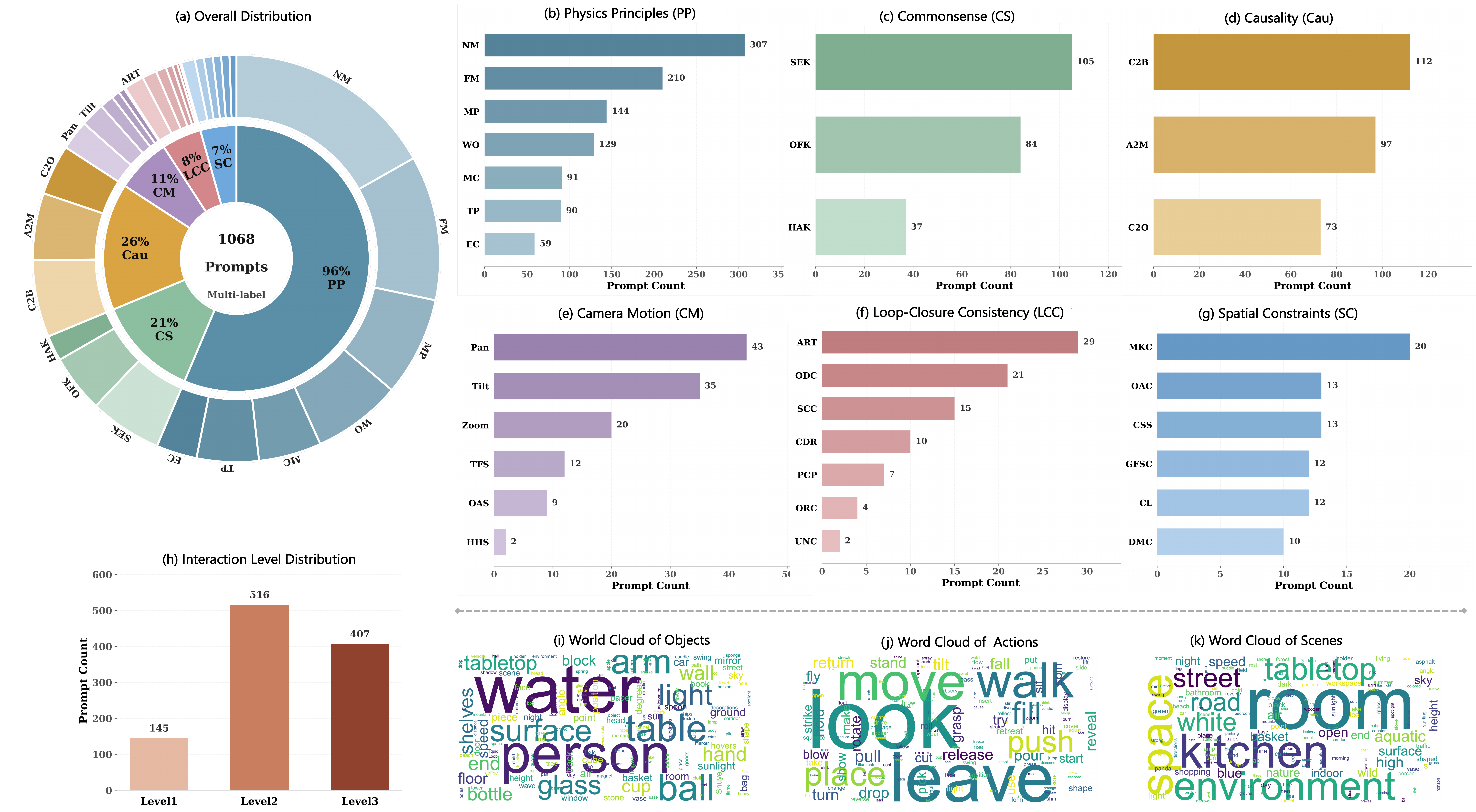}
    \caption{Statistics of Omni-WorldSuite. 
    \textbf{(a)} Overall Distributions; \textbf{(b--g)} Distributions of core principles; \textbf{(h)} prompt counts by interaction level; \textbf{(i--k)} word clouds of objects, actions, and scenes. NM (Newtonian Mechanics), FM (Fluid Mechanics), MP (Material Properties), WO (Waves and Optics), MC (Momentum and Collision), TP (Thermodynamics and Phase Transition), EC (Energy Conversion and Conservation); SEK (Scene/Event Knowledge), OFK (Object Function Knowledge), HAK (Human Action Knowledge); C2B (Condition-to-Behavior), A2M (Action-to-Motion), C2O (Collision-to-Outcome); TFS (Tracking / Follow Shot), OAS (Orbit / Arc Shot), HHS (Handheld / Shaky); ART (Axial Round-Trip Motion), ODC (Optical / Dynamic Consistency Closure), SCC (Spiral / Composite Closure), CDR (Curved / Diagonal Return Motion), PCP (Planar Closed-Path Motion), ORC (Orbital / Rotational Closure), UNC (Uncategorized); MKC (Mechanical / Kinematic Constraints), CSS (Contact \& Support Stability), OAC (Occlusion \& Accessibility Constraints), CL (Containment \& Leakage), GFSC (Geometric Fit \& Size Compatibility), DMC (Deformation \& Material Constraints).}
    \label{fig:suite-static}
\end{figure}

\paragraph{Statistics.}
Omni-WorldSuite contains 1,068 evaluation prompts, making it a comparatively large evaluation suite for video generation assessment. As shown in Fig.~\ref{fig:suite-static}(a), the suite exhibits a multi-label distribution over six major annotation dimensions, namely Physics Principles (PP), Commonsense (CS), Causality (Cau), Camera Motion (CM), Loop-Closure Consistency (LCC), and Spatial Constraints (SC). Among these dimensions, Physics Principles appears most frequently, followed by Causality and Commonsense. Fig.~\ref{fig:suite-static}(b--g) further present the subcategory distributions within each dimension. Specifically, NM and FM are the most frequent categories in Physics Principles; SEK dominates the Commonsense dimension; C2B is the most common causal type; Pan and Tilt are the most frequent camera motion patterns; ART and ODC are the most common loop-closure categories; and MKC appears most frequently among the spatial constraints. Fig.~\ref{fig:suite-static}(h) further shows that Level 2 contains the largest number of prompts, followed by Level 3 and Level 1. In addition, the word clouds in Fig.~\ref{fig:suite-static}(i--k) highlight the diversity of objects, actions, and scenes covered by the suite. Overall, these statistics indicate that Omni-WorldSuite is not only large in scale, but also broad in semantic and structural coverage, providing a diverse testbed for evaluating interactive world modeling under physical, causal, spatial, and motion-related constraints.

\section{Omni-Metric}
\label{sec:metrics}

% \begin{figure}
%     \centering
%     \includegraphics[width=1\linewidth]{Images/metrics.png}
%     \caption{Omni-Metrics}
%     \label{fig:metrics}
% \end{figure}

% To systematically evaluate the interactive response capabilities of world models, \textbf{Omni-Metric} introduces three core metrics—Interaction Effect Fidelity, Non-Intervention Consistency, and Spatiotemporal Causal Coherence. These dimensions provide a rational framework for assessing world modeling by measuring the effects of interaction actions on outcomes and state evolution.
% final outcomes and intermediate state evolution.

%方宸v2
To facilitate an \textit{omni}-directional assessment of world models, we introduce \textbf{Omni-Metric}, a framework designed to deliver a truly comprehensive evaluation experience. Omni-Metric delineates three pivotal dimensions: \textit{Generated Video Quality}, which quantifies both static and dynamic visual fidelity; \textit{Camera-Object Controllability}, which scrutinizes scene coherence and object controllability in the absence of external interventions; and \textit{Interaction Effect Fidelity}, which evaluates adherence to physical laws, event interactions, and temporal sequence logic within realistic scenarios. Collectively, these dimensions establish a rigorous paradigm for benchmarking the perceptual quality, environmental stability, and causal reasoning capabilities inherent to advanced world models.

\subsection{Structured Information Extraction}
\label{sec:metrics_struct_infor}
Given a world model to be evaluated, it generates a video $v$ conditioned on an evaluation prompt $P$ (optionally with an initial frame $\text{I}$).
Before computing the metrics, we extract structured representations from $V \in \mathbb{R}^{T \times H \times W}$, where $T$, $H$, and $W$ are the number of frames, height, and width.

\textbf{Entity Trajectories.} 
We employ GroundingDINO and SAM to extract temporally consistent segmentation mask sequences for each entity in the video, denoted as $\{ \mathrm{traj}_k \}_{k=1}^{N}$. Here, $\mathrm{traj}_k$ represents the mask sequence of the $k$-th entity (among $N$ entities), which is treated as its trajectory representation.

\textbf{Optical Flow.} 
We use RAFT to estimate the optical flow field $F$ of the video, capturing regional motion intensity and dynamic variations.

\textbf{Relative Camera Motion.}
Following \cite{li2021generalizing}, we approximate the relative camera motion between consecutive frames using optical flow variations, thereby estimating the corresponding camera motion direction and magnitude.

%方宸v2
\subsection{Generated Video Quality}

This section details the Generated Video Quality dimension of Omni-Metrics. For this evaluation, we leverage established metrics from prior benchmarks: specifically, \textbf{imaging quality}, \textbf{temporal flickering}, \textbf{motion smoothness}, and \textbf{dynamic degree} are sourced from VBench \cite{huang2024vbench}, while \textbf{content alignment} is adopted from WorldScore \cite{duan2025worldscore}. To effectively balance static and dynamic video attributes during assessment, we employ AgenticScore to perform adaptive weight allocation across these indicators. Comprehensive details regarding the AgenticScore mechanism are provided in Section \ref{subsec:agentic_score}.

\subsection{Camera-Object Controllability}
\textit{Camera-Object Controllability} evaluates the coherence of static elements, such as scene layouts and object identities, within generated videos. Specifically, in view-following sequences governed by camera trajectories, this metric assesses whether the scene undergoes anomalous variations or if objects remain strictly consistent with the prompt specifications. Furthermore, acknowledging that scene transitions in generated content often induce camera trajectory discontinuities, we incorporate a dedicated transition assessment module to mitigate evaluation biases arising from such interruptions.
This dimension comprises three independent metrics: \textbf{Camera Control}, \textbf{Object Control}, and \textbf{Transition Detection}. Below is a detailed introduction to each metric.

\paragraph{Camera Control.} To quantitatively analyze camera motion errors in videos, we employ the camera control metric proposed in WorldScore \cite{duan2025worldscore}. This metric evaluates discrepancies in camera trajectories by separately assessing rotational and translational components. These error measurements are subsequently normalized to yield a final score, where a higher value indicates superior performance.

\paragraph{Object Control.} To evaluate object generation consistency, existing approaches (e.g., WorldScore) assess the object control metric by detecting objects in videos using models such as GroundingDino and subsequently performing rule-based matching against the prompt text. However, given the inherent limitations in detection accuracy and the susceptibility of rule-based matching to semantic errors arising from synonymy, we propose an improved formulation for this metric. Specifically, we reframe object control as a direct visual question answering (VQA) problem: given a small set of uniformly sampled frames, a multimodal model is asked whether each target object is present in the video, with a constrained binary response. For a video with object list $\mathcal{O}=\left\{o_i\right\}_{i=1}^K$ we query the model independently for each $o_i$ and obtain binary predictions $\hat{y_{i} \in \{0,1\}}$. The final score is computed as $\frac{1}{K} \sum_{i=1}^K \hat{y}_i$, reflecting the proportion of prompt-specified objects that are visually grounded in the generated content. This formulation eliminates brittle rule-based matching and leverages the semantic robustness of large VLMs to synonyms and compositional cues. In addition, uniform temporal sampling offers a lightweight yet effective summary of the video, providing a practical trade-off between computational cost and coverage of object occurrences.

\paragraph{Transitions Detect.} We determine whether a video contains scene transitions using a content-based scene boundary detector. Specifically, we apply PySceneDetect’s \texttt{ContentDetector} \cite{scenedetect}, which computes frame-to-frame visual dissimilarity (in HSV space) and flags a boundary when the change exceeds a threshold $\tau$, subject to a minimum scene length constraint $L$ (in frames) to suppress spurious detections. Given an input video, we first optionally downsample for efficiency, then perform scene detection to obtain a scene list $\left\{\left(t_i^{\text {start }}, t_i^{\text {end }}\right)\right\}_{i=1}^N$. The number of scenes $N$ provides a direct indicator of transitions (a transition exists if $N > 1$). Consistent with the implementation, we map this to a binary score
\begin{equation}
s_{\text {trans }}= \begin{cases}1, & N=1 \\ 0, & N>1\end{cases}
\end{equation}
so that videos without scene cuts receive a full score, while any detected transition yields zero. This formulation provides a simple and robust assessment of temporal continuity by penalizing scene breaks while remaining computationally lightweight.

\subsection{Interaction Effect Fidelity}
As a core contribution of Omni-Metric, the Interaction Effect Fidelity dimension aims to quantitatively assess challenging aspects of video generation, including long-term content consistency and stability, the causal logical ordering of events, and adherence to the physical laws of the real world. To address these challenges, we propose four comprehensive evaluation metrics: \textbf{InterStab-L}, \textbf{InterStab-N}, \textbf{InterCov}, and \textbf{InterOrder}, which are detailed below.

\paragraph{InterStab-L.} 
To rigorously quantify long-horizon temporal coherence, we introduce InterStab-L, which assesses the consistency of visual content across user-specified temporal revisit pairs $\mathcal{R}=\{(t_a, t_b)\}$. 
Formally, continuous timestamps are discretized to frame indices within the video sequence of length $T$. 
For any frame pair $(i, j)$ corresponding to a revisit pair, we define a composite similarity metric $s(i, j)$ that integrates both low-level structural fidelity and high-level semantic consistency:
\begin{equation}
s(i, j)=\frac{1}{2}\left(\operatorname{SSIM}_{\text {gray }}\left(I_i, I_j\right)+\cos \left(\phi(I_i), \phi(I_j)\right)\right),
\end{equation}
where $\mathrm{SSIM}_{\text{gray}}$ denotes the grayscale Structural Similarity Index~\cite{wang2004image}, and $\phi(\cdot)$ represents a pre-trained vision encoder (e.g., the visual tower of CLIP) that maps frames to semantic feature vectors $f$.  To mitigate the degeneracy of trivial static sequences (where high similarity arises from lack of motion rather than stability), we incorporate a dynamics gating mechanism. Specifically, we evaluate similarity across four canonical anchor intervals spanning the video duration; if the average similarity of these anchors exceeds a static threshold $\tau_{\text{static}}$, the metric is penalized to zero to enforce content dynamics. 
Otherwise, InterStab-L is defined as the mean similarity over the revisit set:
\begin{equation}
\text {InterStab-L}=\frac{1}{|\mathcal{R}|} \sum_{\left(t_a, t_b\right) \in \mathcal{R}} s\left(i(t_a), i(t_b)\right) \cdot \mathbb{I}_{\text{dynamic}},
\end{equation}
where $\mathbb{I}_{\text{dynamic}}$ is the validity indicator derived from the anchor check. 
A higher InterStab-L score reflects robust long-term consistency at designated temporal intervals, balancing structural preservation with semantic stability.
% To address challenging revisit scenarios specifically designed in Omni-WorldBench, we introduce this metric to quantify long-term consistency and stability within such complex settings. Revisit frame pairs in $\mathcal{R}$  refer to video frames that reach the same spatial location at different times. For such pairs, the semantic discrepancy is expected to be minimal. Accordingly, we define InterStab-L :
% \begin{equation}
% \mathrm{InterStab\text{-}L}(s)
% =\frac{1}{|\mathcal{R}|}
% \sum_{(i,j)\in\mathcal{R}}
% D(\hat{o}_i,\hat{o}_j),
% \end{equation}
% where the semantic distance $D(\cdot,\cdot)$ is computed based on $\mathrm{SSIM}$ \cite{wang2004image}.

\paragraph{InterStab-N.} 
% \subsection{Non-Intervention Consistency}
% In addition to \textit{Interaction Effect Fidelity}, which analyzes the effect of interaction actions on target entities, \textit{Non-Intervention Consistency} evaluates whether unaffected regions, including non-target entities and the environment, undergo unintended changes. 

Specifically, InterStab-N is used to assess the stability of non-target regions. 
Given the entity masks extracted in Sec.~\ref{sec:metrics}, removing the target masks yields the non-target spatial region $\mathcal{N}$. We then use the flow magnitudes in these regions over the entire video duration $T$ as a measure of their motion energy:
\begin{equation}
E_{non}(s)=\frac{1}{T}\sum_{t=1}^{T}\frac{1}{|\mathcal{N}|}\sum_{x\in \mathcal{N}}\|\mathrm{Flow}_t(x)\|,
\end{equation}
{where $\mathrm{Flow}_t(x)$ denotes the optical flow vector at location $x$ in frame $t$. The resulting motion energy is then mapped to a bounded stability score:}
\begin{equation}
\mathrm{InterStab\text{-}N}(s)=\exp\Big(-\frac{E_{non}(s)}{\beta \times \min(H,W)}\Big), 
\end{equation}
{where $\beta$ is a scaling factor that, together with the frame resolution, normalizes InterStab-N to $[0,1]$. Higher InterStab-N values indicate greater stability in the non-target regions.}

\paragraph{InterCov.} 
InterCov quantifies object-level causal faithfulness in generated videos by verifying whether interaction-affected entities exhibit semantically consistent responses while unaffected entities maintain temporal stability. This metric complements low-level flow-based coverage with high-level semantic validation, leveraging the reasoning capabilities of Vision-Language Models (VLMs) to assess interaction fidelity. Formally, let $\mathcal{O} = \{o_1,\cdots, o_N\}$ denote the set of target entities subject to causal constraints. We employ a VLM-based semantic verifier to evaluate the video sequence, yielding a binary validity signal $v_o \in \{0,1\}$ for each entity $o \in \mathcal{O}$, where $v_o=1$ indicates that the entity's behavior aligns with the prescribed interaction logic (e.g., dynamic response for affected objects, stationarity for others). The metric is defined as the semantic recall of consistent interactions:
\begin{equation}
\text { InterCov}=\frac{1}{|\mathcal{O}|} \sum_{o \in \mathcal{O}} \mathbb{I}(v_o = 1),
\end{equation}
{where $\mathbb{I}(\cdot)$ is the indicator function. Consequently, InterCov serves as a rigorous measure of object-level semantic consistency, ensuring that generated dynamics adhere to the underlying causal structure.}
{\paragraph{InterOrder.} 
This metric quantifies the alignment between the chronology of propagated events and the ground-truth sequence $\mathcal{E}=\{e_i\}_{i=1}^{K}$. 
Specifically, for any distinct event pair $(e_m, e_n)$ satisfying $m<n$, we employ a pre-trained Vision-Language Model (VLM) as an automated verifier to assess both the occurrence of the events and their relative temporal precedence via a structured query protocol. 
An event pair is deemed \textit{temporally consistent} if the generated sequence preserves the ground-truth ordering. 
Formally, InterOrder is defined as the ratio of consistent event pairs $K_s$ to the total number of possible pairs:}
\begin{equation}
\mathrm{InterOrder} = \frac{2K_s}{K(K-1)},
\end{equation}
where $\mathrm{InterOrder} \in [0,1]$. 
A higher score indicates superior capability in maintaining temporal coherence and logical event progression.

\subsection{AgenticScore}
\label{subsec:agentic_score}
To accommodate diverse application scenarios and capture different aspects of interactive representation ability, the prompts in Omni-WorldSuite emphasize different evaluation focuses. Therefore, when aggregating the metrics to obtain the final score, each prompt should assign different weights to different evaluation dimensions rather than simply averaging all metrics.
Inspired by agent-based frameworks, we treat each evaluation metric as an independent evaluation agent. Each metric agent first produces a score for its corresponding dimension, after which an aggregation agent adaptively combines these results according to the semantic content of the prompt to produce the final score.

Specifically, the three interaction-centered evaluation agents—interaction effect fidelity $A_I$, generate video quality $A_G$, and camera-object controllability $A_C$—each compute their scores by averaging the results of their respective sub-metrics. 
For example, $A_I = (\mathrm{InterStab-L}+\mathrm{InterStab-N}+\mathrm{InterCov} + \mathrm{InterOrder}) / 4$.
The aggregation agent then analyzes the relative importance of these three evaluation dimensions using an MLLM conditioned on the evaluation prompt, and maps the resulting ranking to predefined weight coefficients $w_1, w_2, w_3$.

% In addition, following WorldScore, we introduce a general visual quality agent $A_C$, which is computed as the average of four sub-metrics: ClipScore, Motion Smoothness, Optical Flow Consistency, and Object Detection Accuracy, with weight coefficient $w_C$.
The final score, \textbf{AgenticScore}, is defined as:
\begin{equation}
\mathrm{AgenticScore}
= w_1 A_I + w_2 A_G + w_3 A_C .
\end{equation}
\section{Experiments}
\label{sec:experiments}

\subsection{Models and Evaluation Protocol}

\paragraph{Evaluated Models.}
Across distinct generation tasks—namely Text-to-Video (T2V; Director3D~\cite{wu2024direct3dscalableimageto3dgeneration}, OpenSoraPlan~\cite{lin2024opensoraplan}, T2V-Turbo~\cite{li2024t2v-turbo}, HunyuanVideo~\cite{kong2024HunyuanVideo}), Image-to-Video (IT2V; Matrix Game2.0~\cite{he2025matrix}, Wan2.1~\cite{wan2025}, Wan2.2~\cite{wan2025}, CogVideo~\cite{hong2022CogVideo}, OpenSora~\cite{zheng2024opensora}, Cosmos~\cite{agarwal2025Cosmos}, LargeVideoPlanner~\cite{chen2025large}), and camera-controlled generation (HunyuanWorld~\cite{hyworld2025}, HunyuanGameCraft~\cite{li2025hunyuan}, ViewCrafter~\cite{yu2024viewcrafter}, Gen3C~\cite{ren2025gen3c}, Lingbot~\cite{team2026advancing}, FantasyWorld~\cite{dai2025fantasyworldgeometryconsistentworldmodeling}, WonderWorld~\cite{yu2025wonderworld})--we evaluate a total of 18 representative world models encompassing diffusion-based, autoregressive, and hybrid paradigms.

\paragraph{Evaluation Protocol.}
We comprehensively evaluate the generative capabilities of world models using our proposed benchmark, \textbf{Omni-WorldBench}. The evaluation protocol is driven by \textbf{Omni-Metric} (defined in Sec.~\ref{sec:metrics}), which encompasses 15 metrics across three distinct dimensions: (1) generated video quality, (2) interaction effect fidelity, and (3) camera and object controllability. To compute these metrics, we introduce a custom test set, \textbf{Omni-WorldSuite}. Specifically, we evaluate T2V and IT2V models using 410 diverse prompts from this suite, while camera-conditioned models are assessed using a dedicated set of 120 prompts equipped with explicit camera trajectories.

\begin{sidewaystable}[p]
\centering
\caption{Quantitative evaluation results of various models on the proposed benchmark. The metrics are grouped into Interaction Effect Fidelity, Generated Video Quality, Camera-Object Controllability, and the overall AgenticScore. The best results within each group are highlighted in bold. Avg.=average.}
\label{tab:main_results}
\sisetup{
  table-number-alignment=center,
  table-format=3.2,
  detect-weight=true,
  detect-inline-weight=math
}
\small
\setlength{\tabcolsep}{1.5pt}
\renewcommand{\arraystretch}{1.5}
\begin{adjustbox}{width=\textwidth}
\begin{tabular}{
l
*{5}{S}
*{6}{S}
c
*{3}{S}
S
}
\toprule
& \multicolumn{5}{c}{Interaction Effect Fidelity}
& \multicolumn{6}{c}{Generated Video Quality}
& \multicolumn{4}{c}{Camera-Object Controllability}
& \multicolumn{1}{c}{AgenticScore} \\
[2pt]
\cmidrule(lr){2-6}\cmidrule(lr){7-12}\cmidrule(lr){13-16}\cmidrule(lr){17-17}
Model
& {InterStab-L}
& {InterStab-N}
& {InterCov}
& {InterOrder}
& {Avg.}
& {\makecell[c]{Imaging\\Quality}}
& {\makecell[c]{Temporal\\Flickering}}
& {\makecell[c]{Content\\Alignment}}
& {\makecell[c]{Motion\\Smoothness}}
& {\makecell[c]{Dynamic\\Degree}}
& {Avg.}
& {\makecell[c]{Camera\\Control}}
& {\makecell[c]{Transitions\\Detect}}
& {\makecell[c]{Object\\Control}}
& {Avg.}
& {(\%$\uparrow$)} \\
\midrule

\rowcolor{gray!12}\multicolumn{17}{c}{\textbf{T2V}}\\
Director3D      & 73.49 & 89.24 & 44.48 & 38.41 & 61.41 & 48.90 & \textbf{99.48} & 89.87 & \textbf{99.68} & 47.31 & 77.05 & \textemdash & \textbf{99.75} & 72.07 & 85.91 & 71.00 \\
OpenSoraPlan    & 68.78 & \textbf{95.76} & 40.29 & 36.59 & 60.36 & 53.55 & 98.67 & 82.16 & 99.22 & 16.83 & 70.09 & \textemdash & 98.29 & 70.65 & 84.47 & 68.10 \\
T2V-Turbo       & \textbf{82.98} & 66.18 & 43.83 & 36.70 & 57.42 & \textbf{63.48} & 97.64 & \textbf{90.24} & 98.54 & \textbf{47.80} & \textbf{79.54} & \textemdash & 99.02 & 73.75 & 86.39 & 69.85 \\
HunyuanVideo    & 77.35 & 82.37 & \textbf{53.02} & \textbf{46.78} & \textbf{64.88} & 61.60 & 98.67 & 81.18 & 99.31 & 44.88 & 77.13 & \textemdash & 98.54 & \textbf{85.30} & \textbf{91.92} & \textbf{73.96} \\

\rowcolor{gray!12}\multicolumn{17}{c}{\textbf{IT2V}}\\
Matrix Game2.0       & 47.38 & 19.96 & 55.27 & 48.41 & 42.76 & 58.85 & 95.37 & 62.44 & 98.17 & \textbf{99.02} & 82.77 & \textemdash & 53.41 & 87.85 & 70.63 & 60.33 \\
Wan2.1             & 70.98 & 58.53 & \textbf{64.52} & \textbf{54.19} & 62.06 & 65.89 & 96.75 & 81.56 & 98.04 & 70.98 & 82.64 & \textemdash & 81.95 & \textbf{91.93} & 86.94 & 73.21 \\
Wan2.2             & 79.68 & 79.98 & 56.99 & 52.70 & \textbf{67.34} & \textbf{66.83} & 98.36 & 79.67 & 99.09 & 46.83 & 78.16 & \textemdash & 96.83 & 91.18 & 94.01 & \textbf{75.92} \\
CogVideo           & 79.03 & 79.51 & 54.80 & 48.98 & 65.58 & 61.47 & 98.04 & 79.19 & 98.84 & 29.02 & 73.31 & \textemdash & 97.56 & 87.33 & 92.45 & 73.27 \\
OpenSora           & 66.68 & 69.90 & 62.54 & 48.17 & 61.82 & 57.40 & 98.29 & \textbf{86.09} & 99.09 & 79.76 & \textbf{84.13} & \textemdash & 95.12 & 90.51 & 92.82 & 74.71 \\
Cosmos             & 79.55 & 79.63 & 53.89 & 51.81 & 66.22 & 66.30 & 98.29 & 80.93 & 99.17 & 44.15 & 77.77 & \textemdash & \textbf{98.78} & 91.01 & \textbf{94.90} & 75.42 \\
LargeVideoPlanner  & \textbf{82.15} & \textbf{87.43} & 42.84 & 45.15 & 64.39 & 66.60 & \textbf{98.99} & 77.67 & \textbf{99.36} & 32.44 & 75.01 & \textemdash & 97.32 & 89.84 & 93.58 & 73.42 \\

\rowcolor{gray!12}\multicolumn{17}{c}{\textbf{With Camera}}\\
HunyuanWorld        & 77.49 & \textbf{67.92} & 55.31 & 48.15 & \textbf{62.22} & 64.27 & \textbf{97.65} & \textbf{76.10} & \textbf{99.14} & 67.32 & 80.90 & 55.40 & 96.10 & 87.52 & 79.67 & \textbf{74.36} \\
HunyuanGameCraft  & 64.78 & 51.28 & 46.74 & 37.50 & 50.08 & 67.09 & 96.12 & 47.29 & 98.67 & 91.67 & 80.17 & 27.96 & 95.00 & 84.55 & 69.17 & 67.39 \\
ViewCrafter         & 81.15 & 4.22 & 43.19 & 41.11 & 42.42 & 61.37 & 91.01 & 49.03 & 95.40 & \textbf{100.00} & 79.36 & 42.91 & 95.00 & 86.17 & 74.69 & 65.88 \\
Gen3c               & 75.90 & 38.40 & \textbf{57.50} & \textbf{53.75} & 56.39 & 58.55 & 95.78 & 63.84 & 98.86 & 98.33 & 83.07 & 48.07 & 85.83 & 84.55 & 72.82 & 71.61 \\
Lingbot             & 74.84 & 66.59 & 45.28 & 35.28 & 55.50 & \textbf{67.65} & 96.93 & 52.83 & 98.67 & 45.83 & 72.38 & 33.97 & \textbf{98.33} & 89.76 & 74.02& 67.16 \\
FantasyWorld        & 72.66 & 55.34 & 48.40 & 41.94 & 54.59 & 64.87 & 96.32 & 56.32 & 98.68 & 73.33 & 77.90 & 42.29 & 93.33 & \textbf{90.45} & 75.36 & 69.49 \\
WonderWorld         & \textbf{84.96} & 24.89 & 51.26 & 43.84 & 51.24 & 60.40 & 92.26 & 74.22 & 99.02 & \textbf{100.00} & \textbf{85.18} & \textbf{96.12} & 73.95 & 87.33 & \textbf{85.80} & 74.02 \\
\bottomrule
\end{tabular}
\end{adjustbox}
\end{sidewaystable}

\subsection{Implementation Details}

All inference experiments are conducted using NVIDIA H20 GPUs. To ensure optimal performance and fair comparison, the software environments—specifically the Python and PyTorch versions—are strictly configured according to the official guidelines provided by each model's respective codebase. 

\paragraph{Text-to-Video (T2V) Models.}
For the T2V generation paradigm, models are conditioned solely on text prompts. Specifically, HunyuanVideo generates 91 frames at a $1280\times 720$ resolution using 50 inference steps at 16 FPS. OpenSoraPlan (v1.0.0) employs the T5-v1.1-XXL text encoder, producing 65 frames at $512\times 512$ resolution with 250 sampling steps and a classifier-free guidance (CFG) scale of 7.5 at 24 FPS. T2V-Turbo (v2-no-MG V) generates 40 frames at 8 FPS, utilizing 32 inference steps and a CFG scale of 7.5. Notably, Director3D relies on its self-predicted camera trajectories for novel view rendering, outputting $960\times 960$ resolution videos.

\paragraph{Image-to-Video (IT2V) Models.}
IT2V models utilize both a starting frame and text prompts as conditioning inputs. Both Wan2.1 (14B-720P) and Wan2.2 (A14B) generate 81 frames (5 seconds) at $1280\times 720$ resolution operating at 16 FPS; however, Wan2.1 uses 50 steps with a 5.0 guidance scale, whereas Wan2.2 uses 40 steps with a 3.5 guidance scale. Cosmos (Cosmos-predict-14B) operates at the same resolution and frame rate but outputs 77 frames using 35 steps and a guidance scale of 7. CogVideo (CogVideoX-5b-I2V) generates 49 frames at $720\times 480$ (8 FPS, 50 steps, CFG scale 6). OpenSora (v2) is configured to a 256px (16:9) resolution, yielding 129 frames at 24 FPS with 50 steps and a 7.5 CFG scale. LargeVideoPlanner leverages a base model for $832\times 480$ resolution (81 frames, 16 FPS, 40 steps) with customized history and language guidance scales of 1.5 and 2.5, respectively. Finally, Matrix-Game2.0 (universal mode) outputs $650\times 352$ videos at 16 FPS, relying on randomly generated camera trajectories.

\paragraph{Camera-Conditioned Models.}
To evaluate camera controllability, these models require explicit camera parameters or trajectories. HunyuanWorld (v1.5, Autoregressive-480P-I2V) generates videos at $800\times 496$ resolution and 16 FPS. Hunyuan-GameCraft utilizes complete pose information to generate 132 frames at $704\times 1216$ (24 FPS). ViewCrafter adopts an equidistant camera pose sampling strategy, producing 25 frames at $576\times 1024$ (8 FPS). Both Gen3C and Lingbot operate at $720\times 1280$ resolution, with their outputs consistently truncated to the first 121 frames. Furthermore, FantasyWorld ($832\times 480$) and WonderWorld ($512\times 512$) employ a frame-subsampling strategy, compressing the original 132-frame camera trajectories down to 81 frames. It is worth noting that for WonderWorld, large-scale camera motions in the dataset may occasionally result in blank frames during rendering due to incomplete point cloud coverage.

\begin{figure}
    \centering
    \includegraphics[width=\linewidth]{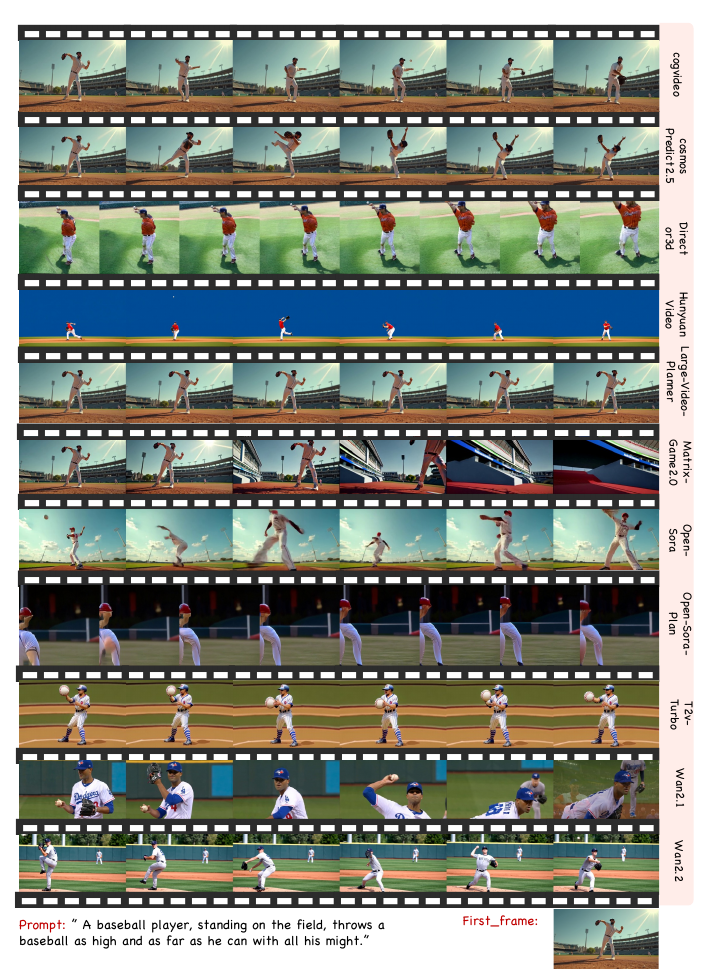}
    \caption{\textbf{Non-camera-controlled Interaction Comparison.} Qualitative comparison of generated videos from different models under the same prompt and first-frame condition. Representative frames illustrate differences in interaction effect fidelity, motion dynamics, and scene coherence during the throwing action.}
    \label{fig:visual}
\end{figure}

\begin{figure}
    \centering
    \includegraphics[width=\linewidth]{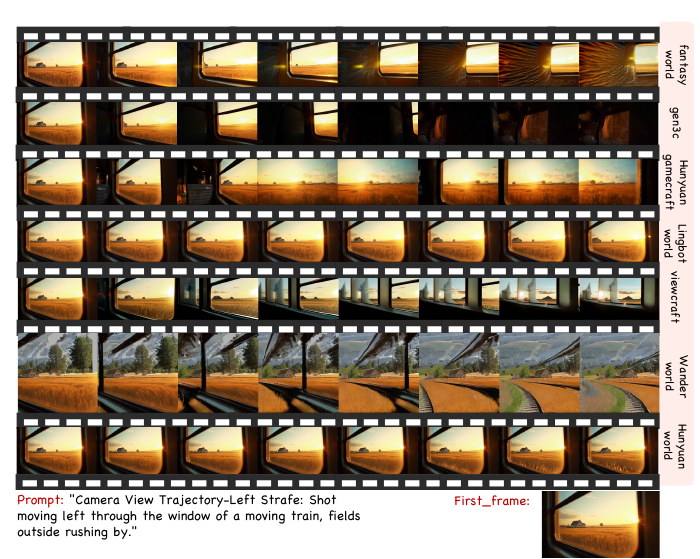}
    \caption{\textbf{Camera-Controlled Interaction Comparison.} Qualitative Comparison of Generated Videos from Different Models under the Same Prompt, First-Frame, and Camera Trajectory Condition.}
    \label{fig:visual2}
\end{figure}

\subsection{Quantitative Evaluation Results and Analysis}
\label{subsec:Omni-Metric_exp}

This section presents a comprehensive automatic evaluation of various advanced video generation models on the proposed benchmark. As shown in Tab.~\ref{tab:main_results}, different model categories exhibit clear trade-offs among interaction fidelity, video quality, and controllability.

\textbf{Overall Performance:} Overall, the Image-to-Video (IT2V) paradigm, which incorporates richer conditional inputs like images, demonstrates the highest performance potential on the current benchmark. Wan2.2 achieves the highest overall AgenticScore across all models at 75.92\%, closely followed by Cosmos (75.42\%). Among pure Text-to-Video (T2V) models, HunyuanVideo performs the best, reaching 73.96\%. In the group supporting explicit camera control (With Camera), hunyuanworld (74.36\%) and wonderworld (74.02\%) take the lead.

\textbf{Interaction Effect Fidelity:} This dimension evaluates the stability of models in handling complex physical and logical interactions. The IT2V group shows high consistency, with Wan2.2 achieving the highest average score of 67.34\%. Notably, some models in the ``With Camera'' group exhibit a significant trade-off across different interaction sub-metrics. For instance, wonderworld scores an impressive 84.96\% on InterStab-L but drops sharply to 24.89\% on InterStab-N. This indicates that maintaining consistent underlying interaction logic while introducing complex camera scheduling remains a challenge for current models.

\textbf{Generated Video Quality:} In terms of basic visual quality, the vast majority of evaluated models have reached extremely high levels in Temporal Flickering and Motion Smoothness (mostly exceeding 95.00\%). However, there is a substantial variance in the Dynamic Degree across models, which constitutes a core differentiator in generation capabilities. ViewCrafter and WonderWorld achieve a perfect score of 100.00\%, while other models in the same group vary significantly. Therefore, the major differences across models no longer mainly come from temporal smoothness, but rather from content alignment and dynamic responsiveness.

This metric directly reflects the models' ability to precisely control specific elements. Camera-aware methods show clear advantages here. WonderWorld demonstrates an overwhelming advantage with an explicit Camera Control score of 96.12\%, far surpassing other models in the same category. Meanwhile, HunyuanWorld obtains the best average controllability score of 79.67\% in its group. Furthermore, regarding Object Control, Cosmos (94.90\%) and Wan2.2 (94.01\%) excel in the IT2V group.

\textbf{Summary:} Current models are already strong in conventional video quality metrics, but still show clear limitations in action-conditioned world evolution, causal interaction consistency, and joint camera-object control. These results highlight the importance of evaluating world models beyond passive video quality and toward agent-centric interactive generation.

\subsection{Qualitative Evaluation}

\paragraph{Visual Comparison of T2V and IT2V Models.} To provide a concrete illustration of our evaluation on interaction effect fidelity and motion dynamics, we present a qualitative comparison in Fig.~\ref{fig:visual}. The models are evaluated under a challenging \textbf{Level-2 interaction prompt} that requires generating a baseball player performing a powerful throw. As shown in the visual sequences, Wan2.2~\cite{wan2025} demonstrates superior performance in this scenario; it successfully synthesizes a complete, anatomically reasonable pitching motion while maintaining the athlete's structural integrity and scene coherence throughout the video. In stark contrast, Matrix-Game2.0~\cite{he2025matrix} struggles significantly with this complex physical interaction. The generated action is not only incomplete but also suffers from severe temporal degradation, culminating in the catastrophic collapse and complete disappearance of the human figure in the final frames. These qualitative observations—particularly the stark disparities in physical interaction and temporal consistency—are highly consistent with the quantitative results presented in Sec.~\ref{subsec:Omni-Metric_exp}, further validating the effectiveness of our Omni-Metric evaluation framework.

\paragraph{Visual Comparison of Camera-Conditioned Models.} In our qualitative analysis, we categorize this example as a \textbf{Level-1 interaction} (camera view trajectory control: left strafe). As shown in Fig.~\ref{fig:visual2}, HunyuanWorld~\cite{hyworld2025} exhibits relatively stable performance throughout the sequence. In contrast, ViewCrafter~\cite{yu2024viewcrafter} introduces a spurious building that appears out of nowhere, degrading visual consistency and leading to a lower score. This qualitative observation is consistent with our quantitative evaluation results presented in Sec.~\ref{subsec:Omni-Metric_exp}, further validating the effectiveness of our Omni-Metric evaluation framework.

\section{Conclusion}
\label{sec:concl}

% In this work, we introduce Omni-WorldBench, a benchmark for evaluating interaction capabilities in video world models. Unlike existing benchmarks that mainly focus on visual quality or motion realism, Omni-WorldBench emphasizes interaction-driven scene evolution and causal consistency under action prompts. To support this goal, we construct Omni-WorldSuite, a prompt suite covering seven physical principles and task-oriented scenarios, organized by interaction complexity. We further propose Omni-Metric, a unified evaluation framework that measures interaction effect fidelity, non-intervention consistency, spatiotemporal causal coherence, and general visual quality, and aggregates them into an overall AgenticScore. Extensive experiments reveal clear differences in their interaction capabilities. While many models achieve strong visual fidelity and motion smoothness, their ability to maintain causal interaction dynamics remains limited. Our results show that Omni-Metric can effectively capture these differences and aligns well with human judgments. We hope Omni-WorldBench can serve as a standardized testbed for advancing interaction-aware video generation and future world models.

\paragraph{Summary.} In this work, we introduce Omni-WorldBench, the first benchmark dedicated to evaluating the interactive response capabilities of video world models. Unlike existing benchmarks that mainly focus on visual quality or motion realism, Omni-WorldBench emphasizes action-driven scene evolution, intermediate state transitions, and causal consistency under interactive prompts, providing a more comprehensive and holistic evaluation perspective. To support this goal, we establish a rigorous evaluation framework consisting of Omni-WorldSuite, a hierarchical prompt suite spanning diverse interaction levels, physical principles, and task-oriented scenarios, and Omni-Metric, an agent-based evaluation protocol that quantitatively measures the impact of actions on both final outcomes and intermediate state transitions, while also assessing non-intervention consistency, spatiotemporal causal coherence, and visual quality, and aggregating them into an overall AgenticScore. Through a systematic evaluation of 18 video generation models and world models, we reveal substantial gaps between visual realism and true interactivity in current systems: although many models achieve strong visual fidelity and motion smoothness, their ability to maintain causally grounded interaction dynamics remains limited. Our results further show that Omni-Metric can effectively capture these differences. We hope Omni-WorldBench can serve as a standardized testbed for diagnosing current limitations and advancing research on more interactive and causally consistent world models, while being continuously refined and extended through community feedback.

\paragraph{Limitations.}
Despite its broad coverage, Omni-WorldBench still has several limitations. Although Omni-WorldSuite spans diverse physical principles, task-oriented scenarios, and interaction levels, it cannot fully capture the complexity of open-world interactive environments, especially long-horizon and highly dynamic settings. In addition, while Omni-Metric provides a unified protocol for evaluating action-conditioned outcomes and intermediate state transitions, we plan to release human-aligned evaluation results in the future to further complement and validate the assessment of interaction quality.

% \newpage
% 注释掉下面的，新建文件.bib
% \section*{References}
{
% \small
\bibliographystyle{IEEEtran}
\bibliography{reference}
}

% References follow the acknowledgments in the camera-ready paper. Use unnumbered first-level heading for
% the references. Any choice of citation style is acceptable as long as you are
% consistent. It is permissible to reduce the font size to \verb+small+ (9 point)
% when listing the references.
% Note that the Reference section does not count towards the page limit.
% \medskip

% {
% \small

% [1] Alexander, J.A.\ \& Mozer, M.C.\ (1995) Template-based algorithms for
% connectionist rule extraction. In G.\ Tesauro, D.S.\ Touretzky and T.K.\ Leen
% (eds.), {\it Advances in Neural Information Processing Systems 7},
% pp.\ 609--616. Cambridge, MA: MIT Press.

% [2] Bower, J.M.\ \& Beeman, D.\ (1995) {\it The Book of GENESIS: Exploring
%   Realistic Neural Models with the GEneral NEural SImulation System.}  New York:
% TELOS/Springer--Verlag.

% [3] Hasselmo, M.E., Schnell, E.\ \& Barkai, E.\ (1995) Dynamics of learning and
% recall at excitatory recurrent synapses and cholinergic modulation in rat
% hippocampal region CA3. {\it Journal of Neuroscience} {\bf 15}(7):5249-5262.
% }

%%%%%%%%%%%%%%%%%%%%%%%%%%%%%%%%%%%%%%%%%%%%%%%%%%%%%%%%%%%%

% \appendix

%%%%%%%%%%%%%%%%%%%%%%%%%%%%%%%%%%%%%%%%%%%%%%%%%%%%%%%%%%%%

\newpage

\end{document}